%% file: main.tex
\documentclass[letterpaper, 10 pt, conference]{ieeeconf}  

\IEEEoverridecommandlockouts                              

\usepackage{graphics} 
\usepackage{epsfig} 
\usepackage{amsmath} 

\usepackage{amssymb}  
\usepackage{pifont}
\usepackage{multirow}
\usepackage{booktabs}
\usepackage[dvipsnames]{xcolor}
\usepackage{tabularx}
\usepackage{comment}
\usepackage{subcaption}
\usepackage{lipsum} 
\usepackage{graphicx,subcaption}
\usepackage[font={footnotesize,it}]{caption}
\usepackage{multirow}

\usepackage{url}

\usepackage{amsmath}
\usepackage{bm}
\usepackage{dsfont}

\usepackage{cite}

\usepackage[ruled,vlined,noend]{algorithm2e}
\usepackage[noend]{algpseudocode}

\usepackage{mathtools} 

\makeatletter
\newcommand{\thickhline}{%
    \noalign {\ifnum 0=`}\fi \hrule height 1pt
    \futurelet \reserved@a \@xhline
}
\newcolumntype{"}{@{\hskip\tabcolsep\vrule width 1pt\hskip\tabcolsep}}
\makeatother

\title{\LARGE \bf
Feel the Tension:  Manipulation of Deformable Linear Objects in  Environments with Fixtures using Force Information}

\author{Finn~Süberkrüb$^1$,
        Rita~Laezza$^2$,
        and~Yiannis~Karayiannidis$^{2,3}$
\thanks{$^1$TC Plattling, Deggendorf Institute of Technology, Germany {\tt\small finn.sueberkrueb@th-deg.de} }
\thanks{$^2$Department of Electrical Engineering, Chalmers University of Technology, Sweden {\tt\small \{laezza , yiannis\}@chalmers.se} }
\thanks{$^3$Department of Automatic Control, Lund University, Sweden {\tt\small  yiannis@control.lth.se} }
\thanks{This work was partially supported by the Wallenberg AI, Autonomous Systems and Software Program (WASP) funded by the Knut and Alice Wallenberg Foundation and the ELLIIT Strategic Area for ICT research, supported by the Swedish Government.}%
}

\usepackage{fancyhdr}
\usepackage{ragged2e}
\usepackage{hyperref}
\fancypagestyle{ieee_notice}
{
    \setlength{\headheight}{32pt}
    \addtolength{\topmargin}{-32pt}
    \lhead{\fbox{\begin{minipage}{\textwidth}
    \noindent\footnotesize\justifying\copyright 2021 IEEE DOI:\href{https://doi.org/10.1109/IROS47612.2022.9982065}{10.1109/IROS47612.2022.9982065}. Personal use of this material is permitted.  Permission from IEEE must be obtained for all other uses, in any current or future media, including reprinting/republishing this material for advertising or promotional purposes, creating new collective works, for resale or redistribution to servers or lists, or reuse of any copyrighted component of this work in other works.\end{minipage}}}
    \cfoot{} 
}

\begin{document}

\maketitle
\thispagestyle{ieee_notice}
\pagestyle{empty}


\begin{abstract}
Humans are able to manipulate Deformable Linear Objects (DLOs) such as cables and wires, with little or no visual information, relying mostly on force sensing. In this work, we propose a reduced DLO model which enables such blind manipulation by keeping the object under tension. Further, an online model estimation procedure is also proposed. A set of elementary sliding and clipping manipulation primitives are defined based on our model. The combination of these primitives allows for more complex motions such as winding of a DLO. The model estimation and manipulation primitives are tested individually but also together in a real-world cable harness production task, using a dual-arm YuMi, thus demonstrating that force-based perception can be sufficient even for such a complex scenario.

\end{abstract}

\section{Introduction}
\label{sec:introduction}
\input{introduction}

\section{Related Work}
\label{sec:related_work}
\input{related_work}

\section{Problem Statement}
\label{sec:problem_statement}
\input{problem_statement}

\section{Model Representation}
\label{sec:model_representation}
\input{model_representation}

\section{Model Estimation}
\label{sec:model_estimation}
\input{model_estimation}

\section{Manipulation Primitives and Control}
\label{sec:manipulation_primitives}
\input{manipulation_primitives}

\section{Experimental Results}
\label{sec:experiments}
\input{experiments}

\section{Conclusion}

We presented a DLO manipulation approach relying only on FT sensing. This is feasible in applications which provide environmental contacts which limit the DLO's DOFs and enable tensioning. By keeping the DLO under tension, it is possible to estimate the location of these contacts, and therefore to keep a simplified path graph representation of the object's state. This representation was successfully employed to solve a cable harness problem. Future work will aim to complement FT sensing with visual information, as well as to equip both grippers with FT sensors which would further improve manipulation.





\bibliographystyle{ieeetr}
\bibliography{references.bib}

\end{document}

%% file: introduction.tex
Force Torque (FT) sensing has been extensively used in robotic contact and interaction tasks both for feedback control schemes and for estimating important properties of the objects being manipulated \cite{RAS22,RAL20,IROS17,Manuelli16}. However, most research has been limited to rigid objects, whose state can be summarized by their pose. Deformable objects constitute a still under-explored class of objects that often require more complex state representations, since such objects may also change in shape. Research on FT-based deformable object manipulation has been limited due to challenges in sensing, since FT measurements require the object to provide sufficient resistance to an applied force. 

This work focuses on Deformable Linear Objects (DLOs), which are characterized by being much larger along one dimension than the other two dimensions  \cite{ASurvey}. Objects like cables, wires and hoses are present in numerous industrial applications, and pose an interesting robotics manipulation challenge. When a DLO with low compression strength such as a rope is manipulated, it does not offer resistance to deformation in any direction, unless it is under tension. Therefore, in the absence of visual information, keeping the DLO taut provides the most amount of information about its state. This property is exploited to formulate our simplified graph representation and nine manipulation primitives. 

\begin{figure}[t]
    \vspace{0.2cm}
    \centering
    \includegraphics[width=\columnwidth]{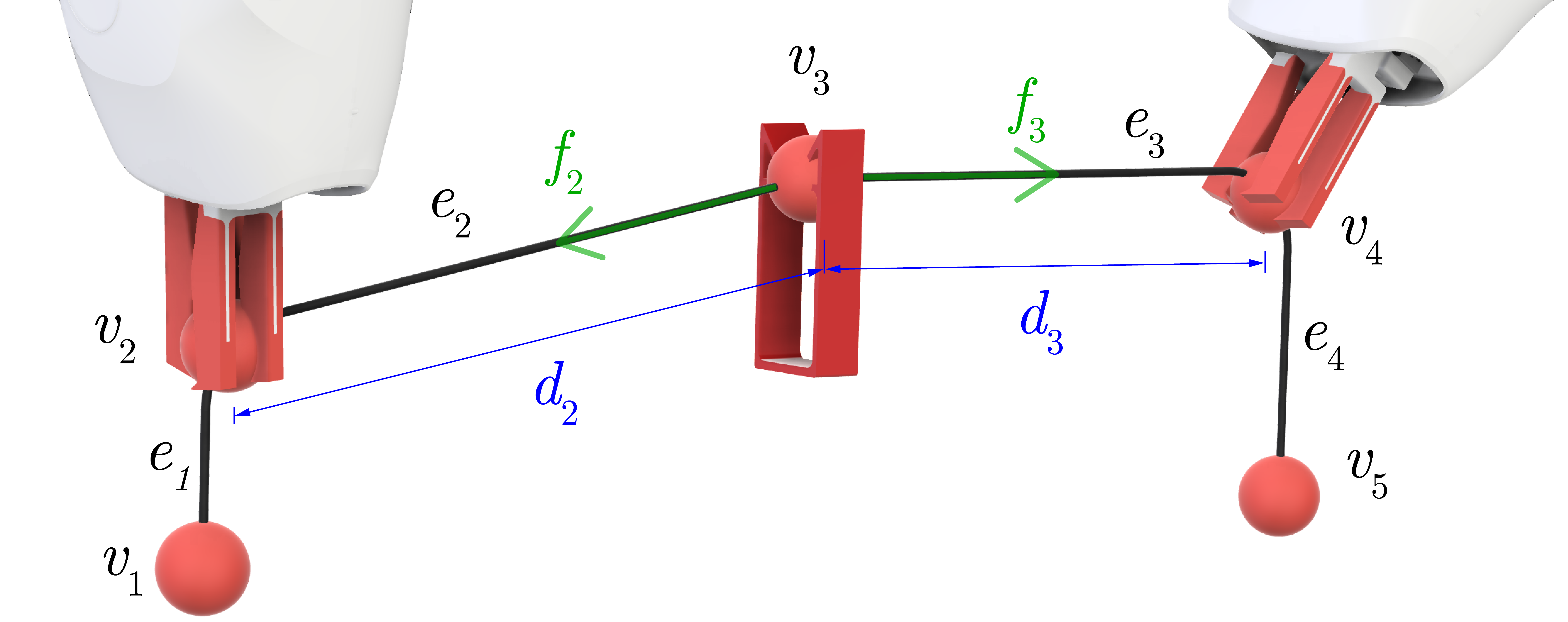}
    \caption{Illustration of DLO model with five feature points $v_1$ to $v_5$, four edges $e_1$ to $e_4$, with forces $f_2$, $f_3$ and distances $d_2$, $d_3$ marked on edges incident on $v_3$. The feature points $v_2$ and $v_3$ are constrained by the YuMi SmartGrippers positions $\bm{p}_{g,right}$ and $\bm{p}_{g,left}$.}
    \label{fig:feature_point_model}
    \vspace{-0.5cm}
\end{figure}

Fig. \ref{fig:feature_point_model} shows an instance of the proposed primitives, namely clipping a DLO into a fixture. To be successful, the DLO must be kept under tension so that it does not bend as it is pushed through the narrow slot. This is an important task for example in wire harness production, where individual cables must be routed through fixtures. Note that due to the challenges of DLO manipulation, manual assembly is still the most common wire harness manufacturing method \cite{9468128}. 

Although visual and tactile sensing are also important for DLO manipulation, in this work we focus on FT information alone. Sanchez et al. \cite{sanchez2020blind} proposed a blind deformable object manipulation approach, but with a more complex model and for a problem without environmental contacts. Here we focus on tasks where the environment provides contacts which constrain the Degrees of Freedom (DOFs) of the DLO. To validate our work, we first evaluate the model estimation in simulation. Secondly, the manipulation primitives are evaluated individually in real-world experiments. Finally, a wire harness problem is solved using a sequence of manipulation primitives. 

%% file: related_work.tex
There has been a growing interest in robotic manipulation of deformable objects, with comprehensive reviews being published on the subject by Sanchez et al. \cite{ASurvey}, Arriola-Rios et al. \cite{A_Tutorial_and_Review}, and Yin et al. \cite{Reviw_Modeling_learning_perception_control}. When addressing deformable object manipulation problems, there are many sub-problems which must be solved. Arriola-Rios et al. divide these into: (i) modeling of shape, (ii) modeling of deformation dynamics, (iii) learning and estimation of model parameters, (iv) perception and prediction, as well as (v) planning and control. Our work addresses elements of all sub-problems.

There are three model types commonly used to represent deformable objects, namely mass-spring systems \cite{SoftMaterialModelingforRoboticManipulation}, position-based dynamics, or finite element methods \cite{Reviw_Modeling_learning_perception_control, StableRealTimeDeformations}. If the object's parameters are known, such approaches can produce realistic models and are already widely used in computer graphics \cite{LearningWhileDoing}. An alternative to physics-based models are purely geometric approaches in which the object is reduced to elementary shapes \cite{doi:10.1177/0278364911430417}. Estimation of model parameters based on sensor data is often necessary.

Visual sensing is the most commonly used modality for perception of deformable objects, since it provides information about overall shape \cite{6630714, SPR}. However, vision may fail in case of occlusions by the robot or by the DLO itself and cannot provide information regarding relatively small deformation and tension. There have been a few works exploring DLO manipulation with environmental contacts based on vision alone \cite{henrich1999manipulating,acker2003manipulating}. Zhu et al. \cite{zhu2019robotic} proposed a motion planning framework, using two motion primitives designed for a robot to shape a DLO through circular contacts. While they use a vision-based contact detector, we attempt to estimate contacts through FT information and make no assumptions on the shape of the contact surface.

Conversely, FT sensing enables the possibility to estimate material properties of an object, such as Young's modulus, but cannot easily be used to identify its global shape. Selective probing can be used in order to combine position information from joint encoders with force information. Estimation of material properties is particularly useful to adapt controller parameters. Sanchez et al. \cite{8594314} presented an approach to estimate the deformation of a foam cube. However, purely force-based position or shape estimators have not yet been used for DLO manipulation problems. FT measurements have also been combined with visual information to estimate material parameters \cite{6263301, 5653949}. 


Regardless of the sensing modality and the DLO model, there are many control strategies in the literature which involve estimating a Jacobian of the object \cite{shetab2022rigid}. For example, Berenson \cite{Manipulation_Without_Modeling_Deformation} introduced the concept of diminishing rigidity to define a Jacobian where points on the DLO closer to the grasped point are assumed to act more rigidly. Other approaches have been proposed where a Jacobian is learned using neural networks \cite{yu2021adaptive} or estimated through weighted least-squares \cite{9133322}. In this work, we instead estimate a reduced graph model. Note that planning of a manipulation sequence is beyond the scope of this paper and other works address that problem \cite{Gabriel, guo2020algorithm}.

%% file: problem_statement.tex
\begin{figure}[tpb]
    \centering
    \includegraphics[width=0.9\columnwidth]{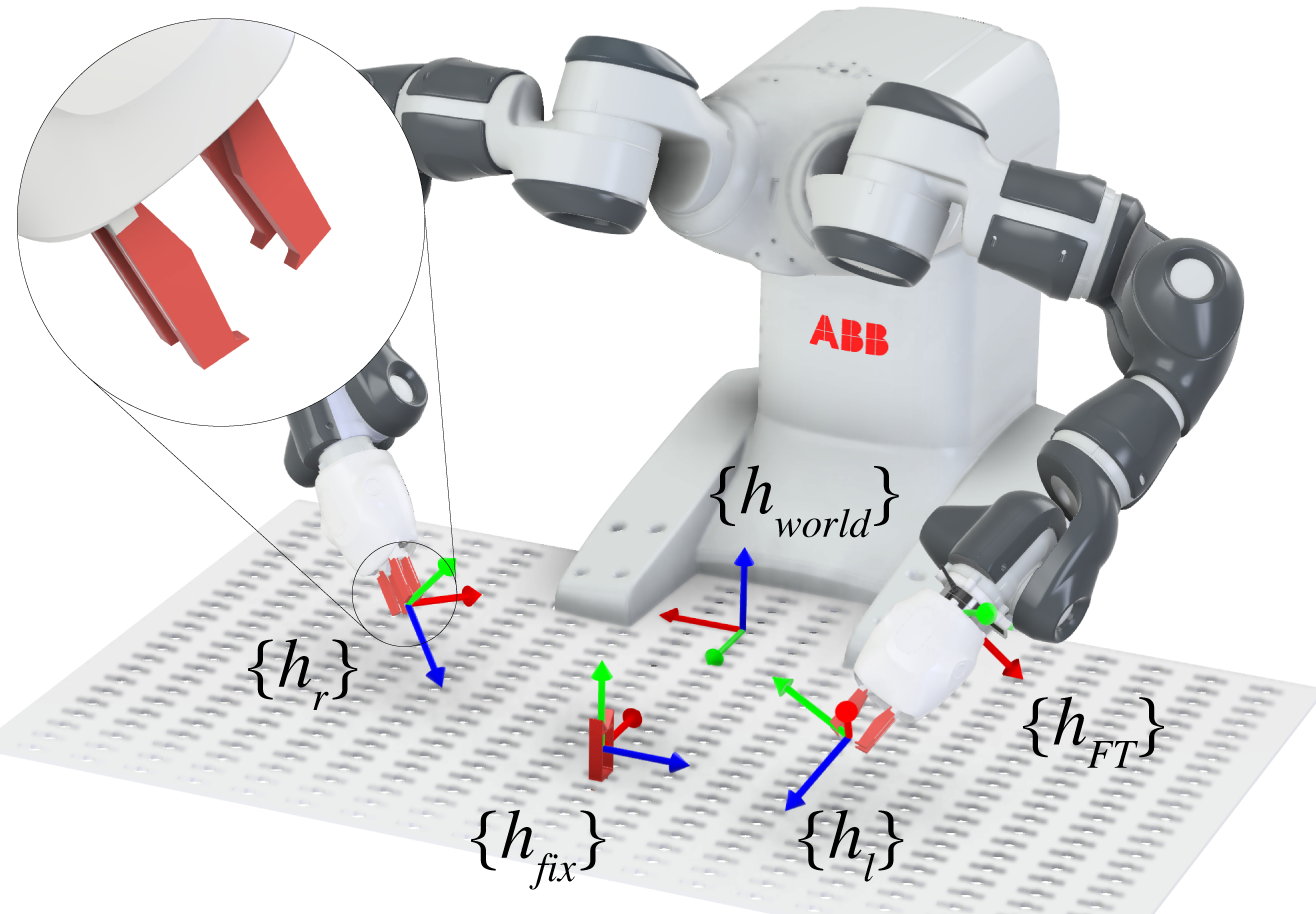}
    \caption{Robotic setup of a YuMi robot equipped with specialized parallel grippers and one fixture attached to a pegboard. Relevant robot frames are indicated as: $\{h_{FT}\}$, the FT sensing frame, $\{h_{l}\}$ the frame associated to the tool tip at the left gripper, and $\{h_{r}\}$ at the right gripper. A fixture frame is denoted as $\{h_{fix}\}$ and the world frame as $\{h_{world}\}$. Red arrows indicate $x$-axes, green arrows $y$-axes and blue arrows $z$-axes. }
    \label{fig:YuMi_frames}
    \vspace{-0.5cm}
\end{figure}

We address force-based manipulation of DLOs without complementary visual information, therefore the robot is assumed to be equipped with at least one wrist FT sensor capable of measuring the tension of the DLO. Furthermore, the problem is formulated assuming that once a DLO is gripped, its local pose is known. Consequently, parallel grippers must prevent involuntary slipping of the DLO along the $x$ (red) and $z$ (blue) axes, while also allowing a controlled sliding motion along $x$ when desired. This can be achieved mechanically through specialized fingertips, shown in Fig. \ref{fig:YuMi_frames}, which guide the DLO to a known location on the fingers and stop any further displacement along $z$.

We assume that the workspace compensates for the blindness of the robot by providing sufficient contact points in the form of fixtures or pivoting points, which restrict the DOFs of the DLO, thus enabling tensioning. This is imperative, since when a DLO is fully stretched between two points, the shape between them can be assumed to be a straight line. Note that gravity itself can be used to restrict the DOFs of the DLO. Once an end-effector equipped with a FT sensor is gripping the object by holding it limp in the air, the shape of the DLO segment below the gripper can be inferred to be vertical, assuming quasi-static motion.

Based on these assumptions, we derive a reduced model of the DLO, presented in Section \ref{sec:model_representation}, which can be estimated online, as described in Section \ref{sec:model_estimation}. This estimated model is then sufficient to define nine elementary manipulation primitives, introduced in Section \ref{sec:manipulation_primitives}.

%% file: model_representation.tex
We propose a reduced model $\{\mathcal{G},\rho\}$ to describe the state of a uniformly weighted DLO based on force-torque information. The state of the DLO is modeled as a path graph $\mathcal{G}=(V,E)$ where: $V$ is a set of feature points on the DLO which can be listed in the order $v_1, v_2, \ldots,v_{|V|}$; and $E$ is a set of edges connecting adjacent feature points, $(v_i, v_{i+1})$ for $i = 1, 2,\ldots, |V| - 1$. This model provides the minimal information necessary to enable a set of elementary manipulation primitives. The model also requires knowledge of the DLO's weight per unit length, $\rho$. If the length and/or weight of the DLO are not known \textit{a priori}, they can be estimated using a sequence of the aforementioned manipulation primitives.  Note that the cardinality of $V$, denoted by $|V|$, is not constant and will change as the manipulator interacts with the DLO and indirectly with the environment. 

\subsection{Feature Points, $V$}
The $i$-th feature point describing the state of the DLO is defined as a tuple $v_i=\left(\mathbf{p}_i, \mathbf{Q}_i, m_i \right)$, where $\mathbf{p}_i \in \mathbb{R}^3$ is the position, $\mathbf{Q}_i \in \mathbb{R}^4$ is the orientation and $m_i \in \mathbb{R}$ is the mass. 

If the feature point is the result of a perceived tension, it may be due to an unknown contact with the environment or another gripper. While in the first case $\mathbf{p}_i$ must be estimated, in the latter it can be obtained directly through forward kinematics. Furthermore, if an additional feature point $v_{k}$ is introduced to define a goal, it can be determined by linear interpolation between two neighboring points, $v_{k-1}$ and $v_{k+1}$. The index $k$ will depend on the location of the added point relative to the existing points in the model. The position is then calculated by interpolation as:
\begin{equation}
    \mathbf{p}_{k} = \mathbf{p}_{k-1} + \delta (\mathbf{p}_{k+1} - \mathbf{p}_{k-1})  \text{, with } \delta \in [0,1]
\end{equation}

Since the orientation cannot be accurately estimated, it is only inferred from the pose of the gripper. For estimated points, the orientation is set so that the $x$-axis is aligned with the length of the DLO, without twist i.e. the orientation of two adjacent points is the same. Similarly to the position, when a point $v_{k}$ is added to describe a goal, a linear interpolation between two quaternion orientations is computed by Slerp (spherical linear interpolation): 
\begin{equation}
    \mathbf{Q}_{k} = \mathbf{Q}_{k-1} ( \mathbf{Q}_{k-1}^{-1} \mathbf{Q}_{k+1})^\delta \text{, with } \delta \in [0,1]
\end{equation}

Finally, the mass of the DLO is equally distributed between the feature points, leading to $m_i$. Depending on the relative location of a feature point with respect to other points, the mass is computed as:
\begin{equation}
    m_i = 
    \begin{cases}
        \frac{{d_i}}{2} \rho & \text{$i$ is an end-point of the DLO}\\
        \frac{d_{i-1}+d_{i}}{2} \rho & \text{$i$ is between feature points}
    \end{cases} \label{eq:mass}
\end{equation}
where ${d_i}$ is a property of the edges to be described next.

\subsection{Edges, $E$}
The $i$-th edge, connecting two adjacent feature points $(v_i,v_{i+1})$ is defined as a tuple $e_i=\left({d_i}, s_i, f_i, k_i \right)$, where ${d_i} \in \mathbb{R}$ describes the length, $s_i \in \mathbb{R}$ is the sag, $f_i \in \mathbb{R}$ the tension and $k_i \in \mathbb{R}$ the spring constant of the DLO between feature point $v_i$ and an adjacent feature point $v_{i+1}$.

\begin{figure}[h!]
    \centering
    \includegraphics[width=\columnwidth]{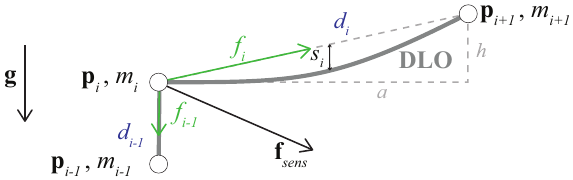}
    \caption{The force measured by an FT sensor in rigid contact with feature point $v_{i}$ is shown as $\mathbf{f}_{sens}$. The position $\mathbf{p}_{i-1}$ and mass $m_{i-1}$ can be determined using knowledge of $\mathbf{p}_{i}$ and the measured force $\mathbf{f}_{sens}$. Force ${f}_{i}$ is a s result of the tension between $\mathbf{p}_{i}$ and $\mathbf{p}_{i+1}$, while the force ${f}_{i-1}$ is caused by the mass $m_{i-1}$. Note that the model assumes ${f}_{i}$ is directed along the straight line. Both forces act on the feature point $v_{i}$.
    Furthermore, the parameters $a$ and $h$ for a catenary curve of a uniformly weighted flexible DLO under tension showing the maximal sag $s_i$.}
    \label{fig:DLO_Force_calculation}
    \vspace{-0.2cm}
\end{figure}

The sag $s_i$ between two feature points, illustrated in Fig. \ref{fig:DLO_Force_calculation}, is a measure used to determine whether the tensioning of the DLO is sufficient for the assumption of a linear connection. If two feature points are not on the same horizontal line and the sag-to-span ratio ($s/a\ll1$) is small enough, no part of the DLO is below the lower feature point. In such cases, the sag can be approximated by equation \eqref{equ:catenarySlack}, defined in \cite{technischeMechanik}:
\begin{equation}
    \label{equ:catenarySlack}
    s_i \approx+ \frac{1}{4}\sqrt{6 \frac{(a^2 + h^2)^2}{a^2} \left[\frac{d_i}{\sqrt{(a^2+h^2)}}-1\right]}
\end{equation}
where $a$ and $h$, shown in Fig. \ref{equ:catenarySlack}, can be computed from the positions of the edge endpoints, namely $\mathbf{p}_i$ and $\mathbf{p}_{i+1}$. The expression is not defined for $a=0$, which occurs when one of the feature points is the end-point of the DLO, constrained by gravity alone. In such cases, the sag is set to $0$. 


The weight compensated DLO tension ${f}_{i}$ between an end-effector at feature point $v_i$ equipped with a FT sensor and a second feature point $v_{i+1}$ can be calculated from the force measurements $\mathbf{f}_{sens}$ and the preceding edge's tension ${f}_{i-1}$ according to:
\begin{equation}
    \hat{\mathbf{p}}_{i} {f}_{i} = \mathbf{f}_{sens} - \mathbf{g}  m_{i} 
+ \hat{\mathbf{p}}_{i-1} {f}_{i-1}
\label{equ:edge_tension}
\end{equation}
where $\mathbf{g} = [0, 0, g]^T$ denotes the gravitational acceleration, and the unit vectors $\hat{\mathbf{p}}_{i-1}$ and $\hat{\mathbf{p}}_{i}$ denote the normalized force directions:
\begin{equation}
    \hat{\mathbf{p}}_{i} = \frac{(\mathbf{p}_{i+1} - \mathbf{p}_{i})}{||\mathbf{p}_{i+1} - \mathbf{p}_{i}||}
\end{equation}


With only one FT sensor, the force ${f}_{i-1}$ can only be taken into account under the assumption that it results from the mass of the rope (a free hanging feature point). In order to sense a second tension e.g. if the DLO is tensioned on both sides of the gripper, another FT sensor would be necessary. For the two terminating feature points only one edge exists and the last part of equation \eqref{equ:edge_tension} is omitted. 

%% file: model_estimation.tex
The model estimation procedure consists of three main parts: (i) estimating the positions of feature points resulting from contact with the environment, (ii) estimating the spring constants of individual edges, (iii) merging estimated features with the known model and interpolate non-estimable feature points between known contact points.


\subsection{Feature Point Estimation}
\label{sec:contactpointestimation}
\begin{figure}[h!]
    \centering
    \includegraphics[width=0.7\columnwidth]{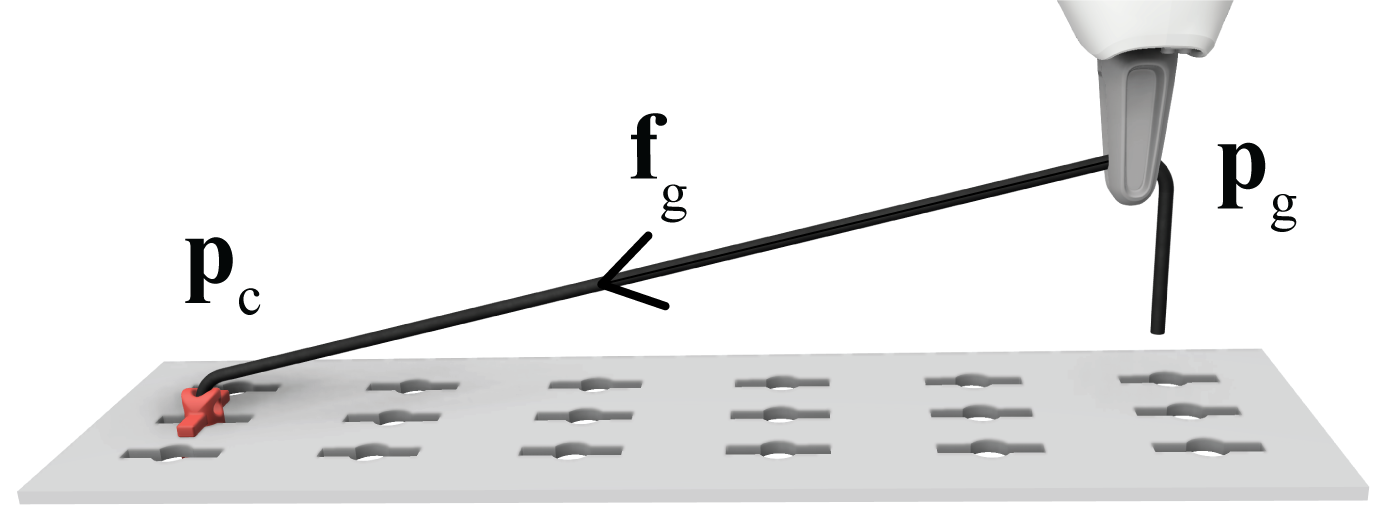}
    \caption{A DLO which is tensioned between the unknown point ${\mathbf{p}}_c$ and the known gripper position ${\mathbf{p}}_{g}$ with the weight compensated force ${\mathbf{f}}_{g}$.}
    \label{fig:Contact_Point_estimation_CAD}
    \vspace{-0.2cm}
\end{figure}

Let us consider the case illustrated in Fig. \ref{fig:Contact_Point_estimation_CAD}, where one gripper is holding the DLO at a position $\mathbf{p}_g$, while the other end is fixed to a single contact, at an unknown position $\mathbf{p}_c$. Both $\mathbf{p}_g$ and $\mathbf{p}_c$ are lying along the line with direction of the tension $\mathbf{f}_{g}$. This leads to the following observation model, for a given measurement $j$:
\begin{equation}
\mathbf{G}_j {\mathbf{p}_{g_j}} = \mathbf{G}_j {\mathbf{p}_{c}}
\label{equ:ls_observation_model}
\end{equation}
where $\mathbf{G}_j$ is the unnormalized projection matrix, which projects the position vectors along the orthogonal complement of the force ${\mathbf{f}_{g}}_j$, defined as:
\begin{equation}
\mathbf{G}_j = {\mathbf{f}^T_{g_j}} {\mathbf{f}_{g_j}}  \mathbf{I}_3 - {\mathbf{f}_{g_j}} {\mathbf{f}^T_{g_j}}
\end{equation}
where $\mathbf{I}_n$ denotes an identity matrix of dimensions $n \times n$.
The contact point position can then be estimated as the intersection of several measurements obtained by changing the gripper position. Equation \eqref{equ:ls_observation_model} can be seen as standard observation model \cite{Intelligente_Verfahren} of the form:
\begin{equation}
    \mathbf{y} = \mathbf{A} \boldsymbol\phi + \mathbf{e}
\label{equ:genneral_leastSquares}
\end{equation}
with  measurement $\mathbf{y} := \mathbf{G} \mathbf{p}_{g}$,  linear mapping $\mathbf{A} := \mathbf{G}$,  parameter $\boldsymbol\phi := \mathbf{p}_c$ and additive noise $\mathbf{e}$.
The same observation model from equation \eqref{equ:ls_observation_model} can be used in a Kalman filter to estimate the contact point $\tilde{\mathbf{p}}_c$. The Kalman update step with a new measurement is given by:
\begin{equation}
   \tilde{\mathbf{p}}_c := {\tilde{\mathbf{p}}_c}^- + \mathbf{K} ( \mathbf{G}_j \mathbf{p}_{g_j} - \mathbf{G}_j{\tilde{\mathbf{p}}_c}^-)
\label{equ:kalmanUpdate}
\end{equation}
where $\mathbf{K}$ is the Kalman gain and ${\tilde{\mathbf{p}}_c}^-$, is the parameter estimated forward in time. The Kalman filter adds the possibility to monitor uncertainties. Certain motion primitives can be executed in order to obtain measurements that improve the estimates, when uncertainty is high.

\subsection{Edge Elasticity Estimation}
An important parameter for force control is the elasticity of the edge. More specifically, it can be used to appropriately tune the tension force controller presented in Section \ref{sec:manipulation_primitives}. Based on Hooke's law, we define:
\begin{equation}
    \text{f}_{g} = k \Delta l
\label{equ:Hookes_law}
\end{equation}
where $\Delta l$ is the displacement of the DLO from its relaxed state and ${\text{f}}_{g}$ the tension which is non-negative. To avoid negative values due to noise, the absolute value of the force is used for the estimation. Further, to avoid dependency on the resting position corresponding to zero tension, we obtain estimates of $k$ by using the backward Euler approximation of the time derivative of equation \eqref{equ:Hookes_law}, that is:
\begin{equation}
  \text{f}_{g}(t)-\text{f}_{g}(t-1)  = k \big[\Delta l(t)-\Delta l(t-1)\big]
\label{equ:spring_measurement} 
\end{equation} 
Compared to the standard observation model from equation \eqref{equ:genneral_leastSquares}, $\text{f}_{g}(t)-\text{f}_{g}(t-1)$ corresponds to the measurement, $k$ to the parameter and $\Delta l(t)-\Delta l(t-1)$ to a regressor.


\subsection{Model Update by Estimates}
\label{sec:constraintmapping}
While a DLO is being manipulated, constraints are dynamically introduced by end-effectors, fixtures, contact points or known distances, and it becomes necessary to update the model accordingly. This is done by mapping the constraints enforced on a feature point to the rest of the model. 
To this end, we define the function $F: \mathcal{C} \times \mathcal{P} \rightarrow \mathcal{P}$, which given current feature point positions in the model $\mathbf{p}_i \in \mathcal{P}$, maps the constrains $\mathcal{C}$ to a new set of positions $\mathcal{P}$. 

Taking the example from Fig. \ref{fig:feature_point_model}, feature points $v_2$ and $v_4$ are fully constrained by a rigid connection to the grippers, while feature points $v_1$ and $v_5$ are constrained by the edge length $d_1$ and $d_4$ and gravity. Furthermore, the feature point $v_3$ was added to the model in order to define a clipping goal. At this stage the position $\textbf{p}_3$ can only be defined by the distances $d_2$ or $d_3$ and linearly interpolated between $v_2$ and $v_4$. In this state, the mapping of the measurements to the model parameters is as follows:

\begin{equation}
\begin{bmatrix}
\mathbf{p}_2 \\
\mathbf{p}_3 \\
\mathbf{p}_4
\end{bmatrix}
=
\left(
\begin{bmatrix}
1 & 0 \\
\frac{d_3}{d_2+d_3} & \frac{d_2}{d_2+d_3}  \\
0 & 1
\end{bmatrix}
\otimes
\mathbf{I}_{3}
\right)
\begin{bmatrix}
\mathbf{p}_{{g, right}}  \\
\mathbf{p}_{{g, left}} 
\end{bmatrix}
\end{equation}
where $\otimes$ denotes the Kronecker product.

As soon as $v_3$ makes contact with the fixture, it is constrained by the estimated contact point. First, the constraint is mapped to the neighbouring points $v_2$ and $v_4$. Then, based on this update, the other points in the model $v_1$ and $v_5$ must also be updated. Since the FT sensor provides only local information about the DLO, knowledge of the previous configuration must be taken into account to determine the global state of the object.


%% file: manipulation_primitives.tex

\begin{figure*}[tbp]
    \centering
    \includegraphics[width=1.0\textwidth]{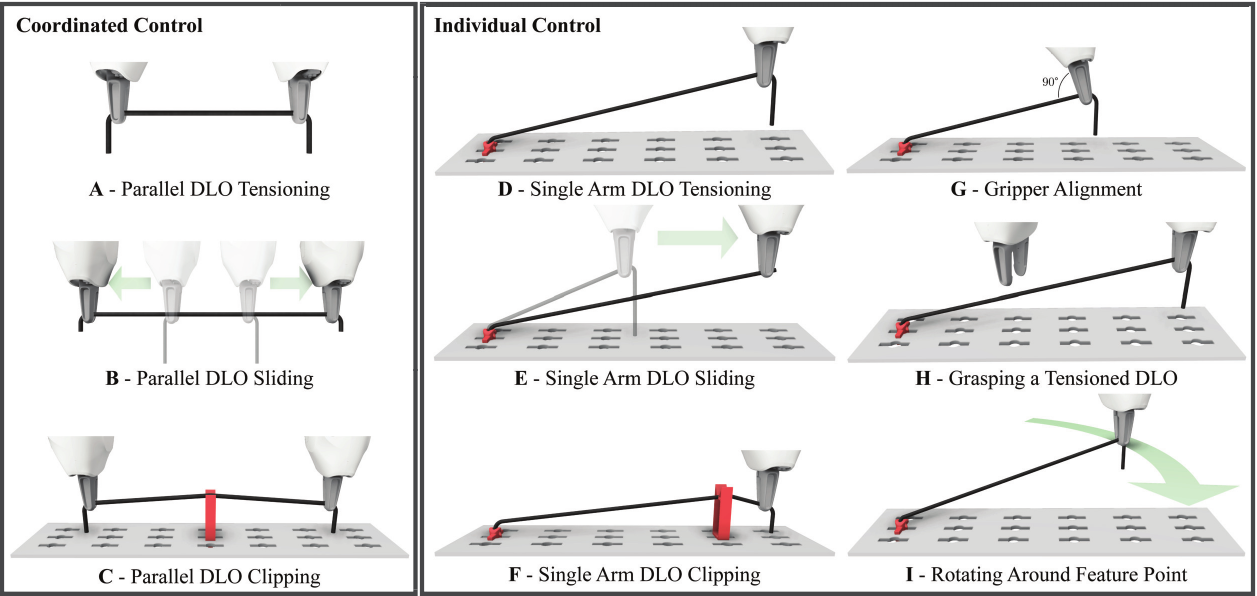}
    \caption{For \textbf{Coordinated Control} the DLO is grasped by two end-effectors, while for \textbf{Individual Control}, the DLO may be grasped by only one end-effector and constrained by a second fixed point e.g. a fixture, another static end-effector, etc. Description of manipulation primitives: 
\textbf{A} - The DLO is tensioned.
\textbf{B} - The DLO is let slide through one of the two grippers in a controlled manner. 
\textbf{C} - The DLO is clamped into a fixture with both grippers in parallel. The fixture exerts a force on the DLO which must be overcome to fasten the DLO properly.
\textbf{D} - The DLO is tensioned between a gripper and a fixed point.
\textbf{E} - The DLO is let slide through the gripper in a controlled manner. 
\textbf{F} - The DLO is clamped into a fixture.
\textbf{G} - The gripper is aligned along the DLO to achieve the most stretched DLO orientation. Any bending will degrade the linear approximation. 
\textbf{H} - The DLO is grasped by another end-effector along its edge.
\textbf{I} - The DLO is rotated around a feature point. This is important to improve estimation or to change the orientation of the DLO under tension.}
    \label{fig:Manipulation_Primitives}
    \vspace{-0.3cm}
\end{figure*}

In this section we present a set of elementary manipulation primitives for DLOs, which can be executed solely based on FT sensing, and minimal knowledge about the environment. These primitives consist of elementary motions such as clipping and sliding, which are typically encountered in wire harness manufacturing. They are illustrated in Fig. \ref{fig:Manipulation_Primitives} together with their description. More complex tasks, such as winding a DLO, are achieved by combining a sequence of primitives.

Each primitive includes a desired force $\text{f}_{d} \in \mathbb{R}$, position $\mathbf{x}_d \in \mathbb{R}^{3}$, quaternion orientation $\mathbf{Q}_d  = [ q_{d_0} \ \mathbf{q}_d ]  \in \mathbb{R}^{4}$ and a projection matrix $\mathbf{P} \in \mathbb{R}^{12 \times 12}$ that specifies the dimensions along which a force control is executed.
The desired end-effector velocities $\boldsymbol\zeta_t$ in task space are provided by a hybrid motion-force controller \cite{Modern_Robotics} given by:
\begin{equation}
\boldsymbol\zeta_t
=
(\mathbf{I}_{12}-\mathbf{P})
\boldsymbol\zeta_m
+
\mathbf{P}
\boldsymbol\zeta_f
\label{equ:HybridMFC_combination}
\end{equation}
where the motion control for both arms is given by  $\boldsymbol\zeta_m \in \mathbb{R}^{12}$ and the force control by $\boldsymbol\zeta_f \in \mathbb{R}^{12}$. The motion control signal $\boldsymbol\zeta_m $ is composed of linear and angular velocity proportional controllers: 
\begin{align}
    \dot{\mathbf{x}}_{m,i} &= k_p(\mathbf{x}_{d,i} - \mathbf{x}_{c,i})\label{equ:Motion_controller_error}\\
\bm{\omega}_{m,i} & = k_o(q_{c_0,i}\mathbf{q}_{d,i} - q_{d_0,i}\mathbf{q}_{c,i}-\mathbf{q}_{d,i} \times \mathbf{q}_{c,i})\label{equ:quaternion_error}
\end{align}
where $k_p$ and $k_o$ are positive control gains, $\mathbf{x}_{c,i}$ and $\mathbf{Q}_{c, i} = [ q_{c_0 ,i} \ \mathbf{q}_{c, i} ]$ describe the current position and orientation, and the index $i$ takes values in the set $\{left,right,abs,rel\}$. Note that the primitives are divided into individual control, where the two end-effectors are controlled separately, with the control signal being defined as:
\begin{align}
    \boldsymbol\zeta_m &= [\dot{\mathbf{x}}_{m,left}^T \quad \bm{\omega}_{m,left}^T\quad \dot{\mathbf{x}}_{m,right}^T \quad \bm{\omega}_{m,right}^T]^T
\end{align}
and coordinated control, where the absolute and relative end-effector position are controlled \cite{Gabriel} using:
\begin{align}
    \boldsymbol\zeta_m &= [\dot{\mathbf{x}}_{m,abs}^T \quad \bm{\omega}_{m,abs}^T\quad \dot{\mathbf{x}}_{m,rel}^T \quad \bm{\omega}_{m,rel}^T]^T
\end{align}

The linear velocity $\dot{x}_{f,i}$ used for force control is obtained by a proportional controller with adaptive damping factor $k_f$:
\begin{equation}
    \dot{x}_{f,i} = k_f(\text{f}_{d,i} - \text{f}_{c,i})
\label{equ:force_controller}
\end{equation}
where $\text{f}_{d,i}$ and $\text{f}_{c,i}$ denote the current and the desired force respectively. By setting $k_f = \hat{k}^{-1}$, where $\hat{k}$ corresponds to the online elasticity estimate, the inverse damping controller in equation \eqref{equ:force_controller} becomes adaptive. 

As a representative example of the tensioning primitives, we present \textbf{D} - \textit{Single arm DLO Tensioning} in more detail. The projection matrix is defined based on the direction between the end-effector and the fixture, that is  $\mathbf{f}_{g} = \mathbf{p}_c - \mathbf{p}_{g}$ (see Fig. \ref{fig:Contact_Point_estimation_CAD}), as follows:
\begin{align}
    \mathbf{P} &= \text{diag}(\mathbf{f}_{g}, \mathbf{0}_3, \mathbf{0}_3, \mathbf{0}_3)
\end{align}
where $\mathbf{0}_n$ denotes a zero vector of dimension $n$ and $\text{diag}(\cdot)$ is a diagonal matrix. Setting $\dot{x}_{f,left}$ as the first three elements of $\boldsymbol\zeta_{f}$, force control is applied uniformly to all three dimensions of the left end-effector. The projection matrix subsequently reduces the force control to the direction along the DLO.
As a representative example of the sliding primitives, we present \textbf{C} - \textit{Parallel DLO Sliding}, where an independent gripping force controller is used:
\begin{equation}
    p_{finger} = k_{f} (\text{f}_{slid}-\text{f}_{c}) + k_i \int (\text{f}_{slid}-\text{f}_{c}) \, dt
\label{equ:Gripping_Force_Controller}
\end{equation}
Here the projection $\mathbf{P}$ is a zero matrix since force control is achieved by adjusting the finger distances $p_{finger}$ based on the frictional tension error $\text{f}_{slid}-\text{f}_{c}$. The integral component, with parameter $k_{i}$, compensates for offset caused by different DLO thicknesses. Furthermore, the motion control uses a constant relative velocity $\dot{x}_{m,rel}$ to slide the end-effector along the DLO while setting $\dot{\mathbf{x}}_{m,abs} :=\mathbf{0}_3$ and $\bm{\omega}_{m,abs} = \bm{\omega}_{m,rel}:=\mathbf{0}_3$.

%% file: experiments.tex
A dual-arm YuMi robot was used for the real-world experiments, where one of the arms was equipped with an ATI Mini40 6-axis force-torque sensor. The environment consisted of a pegboard with DLO mounting possibilities which provided ground truth information about the fixtures. 

\begin{figure*}[tb]
    \begin{minipage}[b]{0.49\textwidth}
        \begin{subfigure}[b]{\textwidth}
            \vspace{0.15cm} 
            \includegraphics[width=\columnwidth]{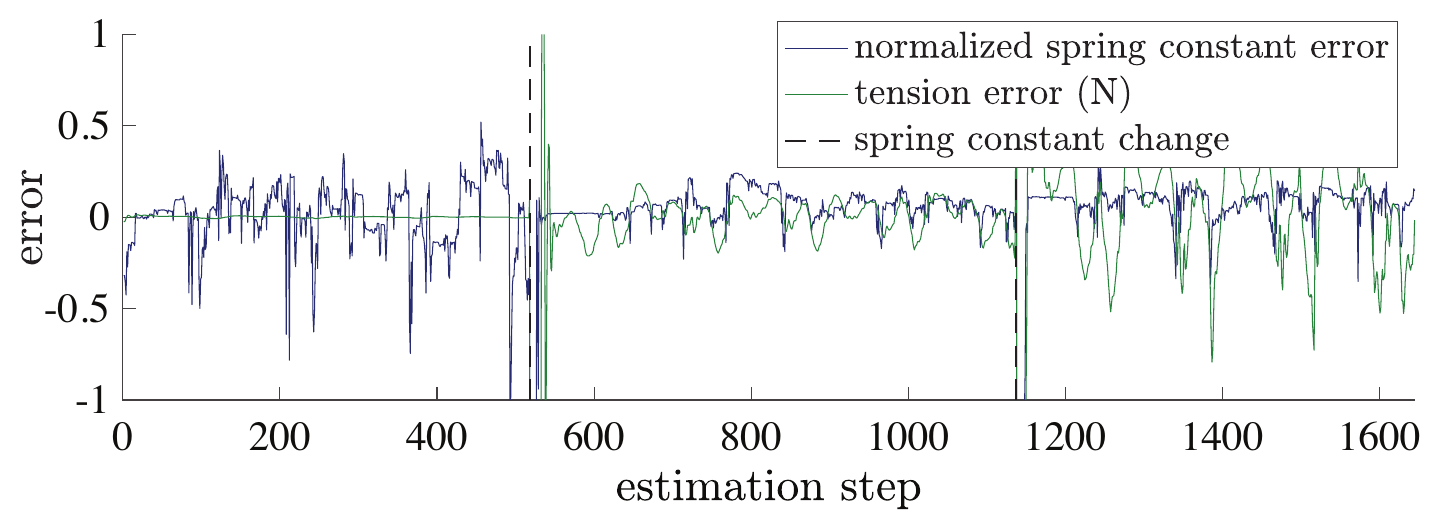}
            \caption{}
            \label{fig:material_parameter_tension_error}
        \end{subfigure}
        \begin{subfigure}[b]{\textwidth}
            \vspace{0.25cm} 
            \centering
            \includegraphics[width=\columnwidth]{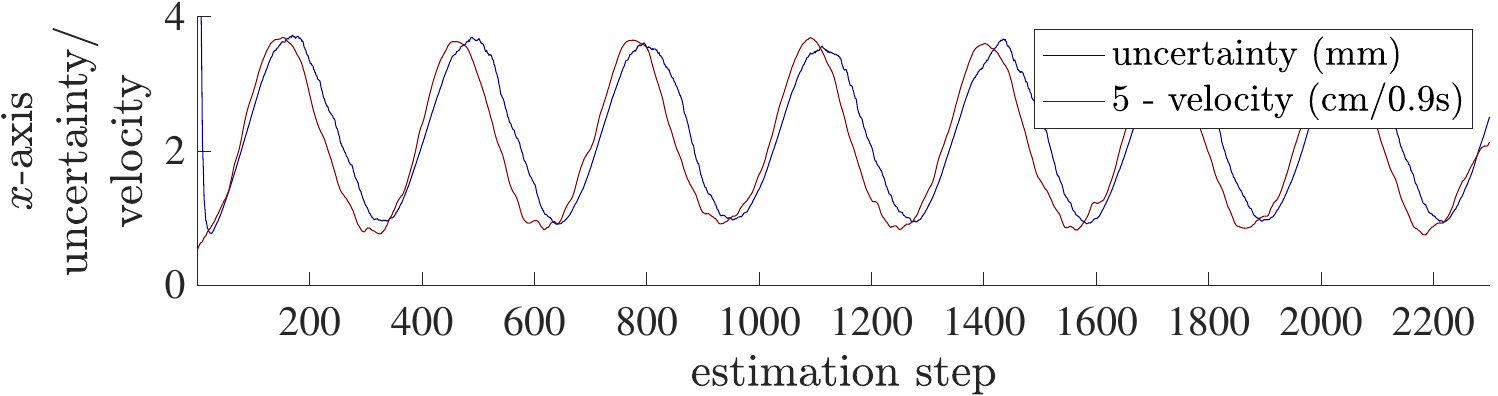}
            \caption{}
            \label{fig:sim_uncertenty_persistence_of_excitation}
        \end{subfigure}
    \end{minipage}
    \begin{subfigure}[b]{.49\textwidth}
        \includegraphics[width=\columnwidth]{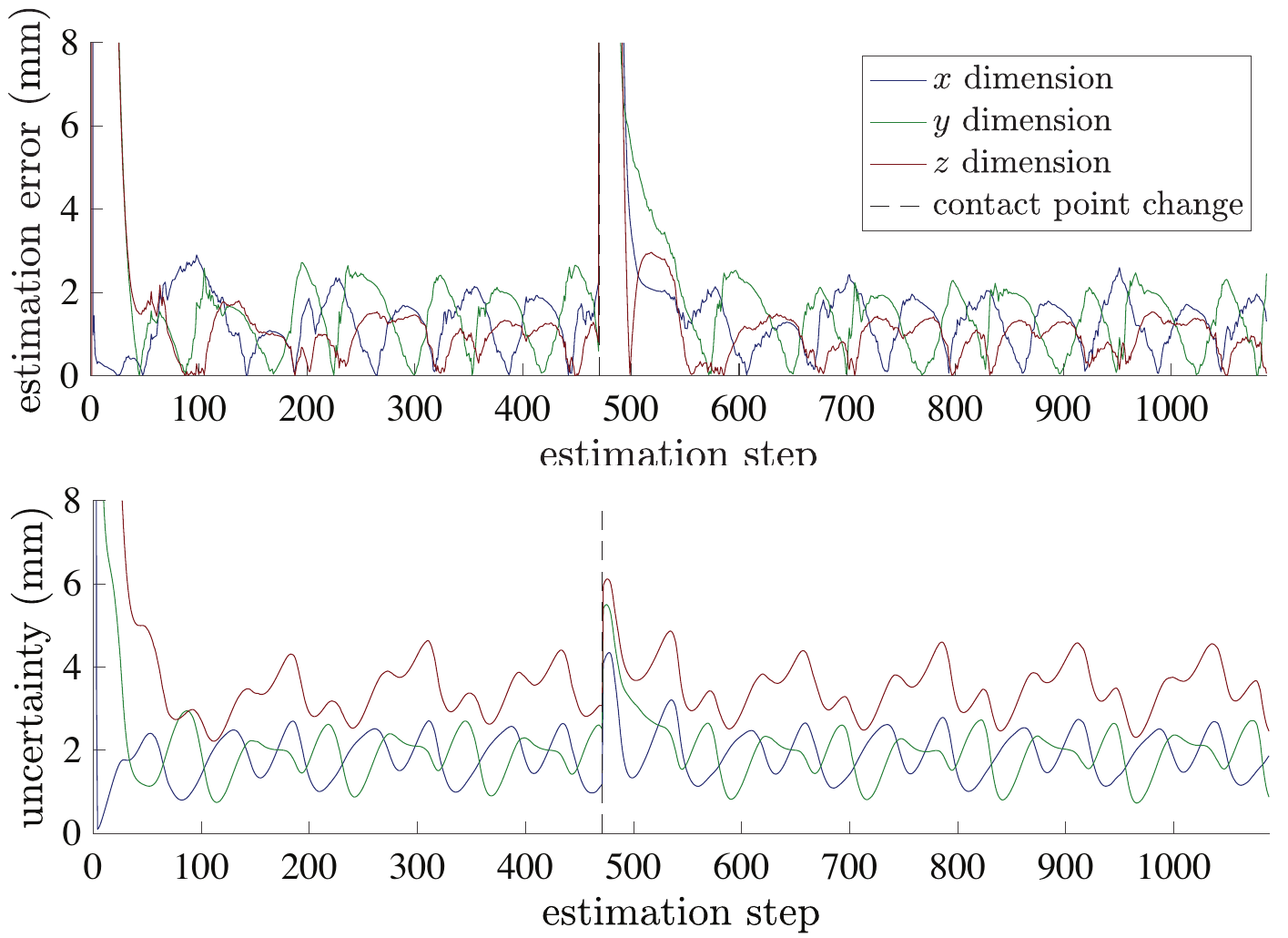}
        \caption{}
        \label{fig:sim_error_uncertenty_Kalman_paper}
    \end{subfigure}
    \caption{Experimental Results: (a) The adaptive damping controller is tested is simulation with three different cables of different elasticities. The plot shows the error as the spring constant is change from $10$ Nm$^{-1}$, to $100$ Nm$^{-1}$ and finally to $1000$ Nm$^{-1}$. (b) Plot of the uncertainty in the position and velocity along the $x$-axis. The velocity is mirrored about the horizontal axis and a positive shift was applied along the vertical axis for visual comparison of the two measurements. (c) Plots of the estimation error (top) and uncertainty (bottom) of the contact point along each dimension. The vertical axis of the top plot is cropped for better visualisation, since the estimation error directly after the contact point change reaches $86$ mm. }
    \label{fig:results}
    \vspace{-0.3cm}
\end{figure*}

\subsection{Elasticity Parameter Estimation and DLO Tensioning}
An essential pre-requisite for the proposed model is to be able to keep the DLO under tension. To evaluate the adaptive controller from equation \eqref{equ:force_controller}, we tested the \textbf{D} -\textit{Single Arm DLO Tensioning} primitive with three different DLO elasticities, as shown in Fig \ref{fig:material_parameter_tension_error}. For all cases the controller was able to maintain the DLOs under tension while keeping all oscillations below the force limits of the robot. Notably, the estimation of the elasticity parameter improves while the tension error increases. A possible explanation can be that large oscillations give a wider range of measurements, thus allowing for a more accurate approximation. On the other hand, a small error leads the system to oscillate only marginally, resulting in very similar measurements. Consequently, the measurement noise has a relatively larger influence, which causes the absolute error to increase.

\subsection{Contact Point Estimation}\label{subsection:Contact_Point_Estimation}

The accuracy of the DLO state representation relies on the identification of contact points by a Kalman filter. In the following experiment, we evaluated the uncertainty of this estimate for manipulation primitive \textbf{I} - \textit{Rotating Around Feature Point}. The distance between the gripper and the fixture is kept constant by a tension controller while a circular motion is executed. The contact point was estimated as described in Section \ref{sec:contactpointestimation}. The uncertainty of the FT sensor was taken from the sensor's specification, while the uncertainty of the end-effector position was disregarded. The process uncertainty ($1 \times 10^{-7}$) was tuned manually in order to achieve a satisfactory trade-off between responsiveness after a contact point change and estimation precision. During the circular motion, the velocity oscillates along the $x$- and $y$-axes. The relationship between the velocity and uncertainty is shown in Fig. \ref{fig:sim_uncertenty_persistence_of_excitation} along the $x$ dimension.

To further test the robustness of the contact point estimation, while executing manipulation primitive \textbf{I} - \textit{Rotating Around Feature Point}, a second fixture was placed in the path of the DLO, thus changing the contact point. Fig. \ref{fig:sim_error_uncertenty_Kalman_paper} consists of two graphs of the same experiment. The top plot shows the position estimation error and the bottom plot shows the uncertainty along the three dimensions. 

Initially, the first contact point was accurately estimated already after half a circle. Note that a full rotation corresponds to approximately 125 iterations of the estimator. Once the contact point changed (at the dashed line), the estimate of the new contact point converged again after an initial spike in the error. For a circular motion, when the velocity along one dimension is higher, measurements with a larger angular difference along this dimension are included in the estimate. However, when there are small displacements along one dimension, very similar measurements are used for the estimation, which means that the measurement error has a relatively larger effect. 

\subsection{Manipulation Primitive Evaluation}
The error rate of the manipulation primitives was evaluated. To this aim all primitives were executed multiple times and the placement of the fixtures, the initial position of the end effectors, as well as the elasticity and length of the DLOs were randomly changed. Results are summarized in Table \ref{tab:evaluation_manipulation_primitives}.
{
\setlength{\tabcolsep}{3pt} 
\renewcommand{\arraystretch}{1.1}
\begin{table}[h]
\centering
\vspace{-0.3cm}
\caption{Repeated experiments to evaluate the reliability of the manipulation primitives, for randomized states.}
\begin{tabular}{lcc}
\thickhline
\textbf{Motion Primitive}                         & \textbf{Repetitions} & \textbf{Failed trials} \\
\thickhline 
Parallel DLO Tension - adaptive control    & 25                   & 2                      \\
Single arm DLO Tension - adaptive control  & 25                   & 1                      \\
Parallel DLO Clipping                      & 15                   & 1                      \\
Single arm DLO Clipping                    & 15                   & 0                      \\
Parallel DLO Sliding                      & 15                   & 0                      \\
Single Arm DLO Sliding                    & 15                   & 0                      \\
Gripper Alignment                          & 15                   & 0                      \\
Grasp a Tensioned DLO - actively tensioned & 15                   & 0                      \\
Rotate Around Feature Point                & 15                   & 0 \\
\thickhline
\end{tabular}
\label{tab:evaluation_manipulation_primitives}
\end{table}}

The three failures of the tensioning primitives were due to poor estimation of the elasticity parameter. At first, the DLO was estimated to be significantly more elastic than it actually was, which caused a movement that quickly built up tension, reaching the limit of the robot. Further, the failure in the \textbf{C} - \textit{Parallel DLO Clipping} primitive was due to an excessive gripper distance which resulted in. The DLO angle upon contact with the fixture was too narrow to apply sufficient force to overcome the clipping force.



\subsection{Harness Production and Model Verification}
A wire harness production scenario was used to demonstrate that manipulating DLOs without visual features, relying on force measurements alone, is feasible. The task required routing two cables through different fixtures, loops, and distances. The target configuration of the cables was predefined as well as the sequence of manipulation primitives\footnote{Video demonstration at \url{https://youtu.be/3Y5HFzrlxds}.}.

\begin{figure}[tbp]
    \centering
    \includegraphics[width=1.0\columnwidth]{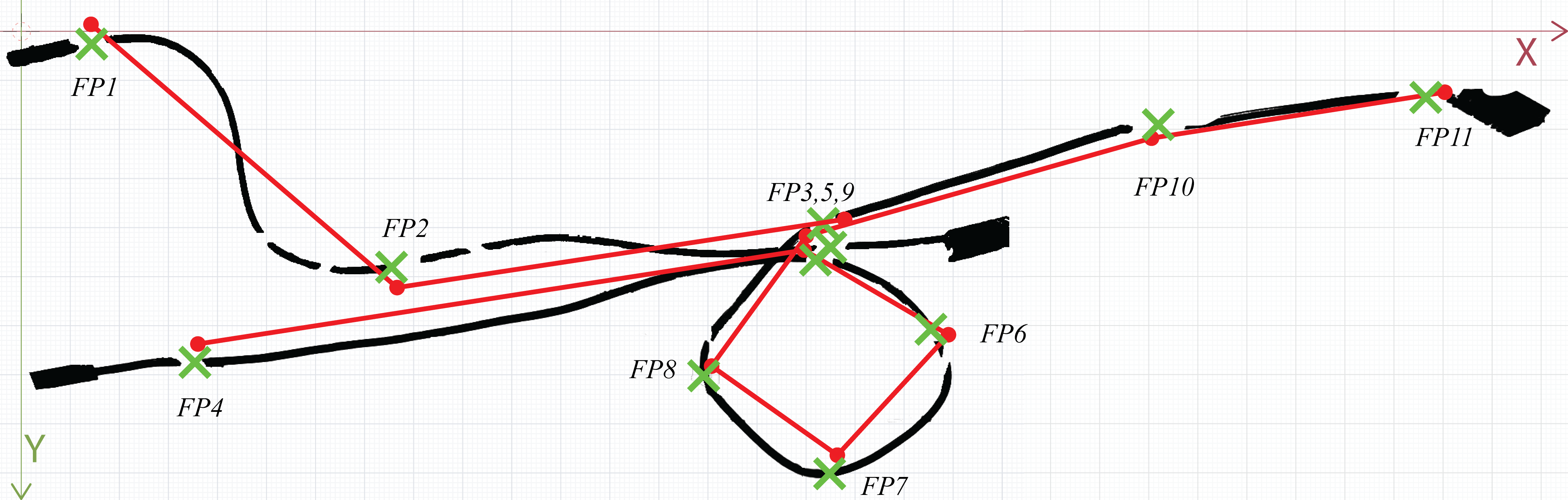}
    \caption{The pixel-based DLO segmentation of the image is shown in black. The feature points and connections of the model are shown in red. The green crosses show the manually selected ground truth contact points.}
    \label{fig:Model_validation}
    \vspace{-0.5cm}
\end{figure}

In the same experiment, the accuracy of the model was validated by determining the ground truth of the routed cable via an external camera. The deviation between model and ground truth in the horizontal plane is on average $7.5$ mm and maximum $14$ mm. Along the vertical $z$-axis average $16$ mm and maximum $28$ mm. The larger deviation along the $z$-axis can be explained by the effect of gravity. The points of the model are estimated when the DLO is under tension. Depending on the fixture, the DLO was displaced by gravity along the $z$-axis before the ground truth data was recorded. The importance of the sag threshold can be understood in the graphical overlay of the segmented image and the DLO model in Fig. \ref{fig:Model_validation}. When segments are not tensioned (top left), the fixed feature points are still correct, but the DLO between the edges is not linear, as assumed in the model. In contrast, for the taut case (top right), the model matches the ground truth data well.

